\documentclass[11pt]{article}
\usepackage{acl2015}
\usepackage{times}
\usepackage{url}
\usepackage{latexsym}
\usepackage{amsmath}

\title{Position-aware Self-attention with Relative Positional Encodings for Slot Filling}

\author{Ivan Bilan \and Benjamin Roth \\
  Center for Information and Language Processing \\
  Ludwig Maximilian University of Munich \\
  Oettingenstr. 67 \\
  Munich, Germany \\
  {\tt ivan.bilan.ua@gmail.com} \\
  {\tt beroth@cis.uni-muenchen.de}\\\
}

\date{}

\begin{document}
\maketitle
\begin{abstract}  
  This paper describes how to apply self-attention with relative positional encodings to the task of relation extraction. We propose to use the self-attention encoder layer together with an additional position-aware attention layer that takes into account positions of the query and the object in the sentence. The self-attention encoder also uses a custom implementation of relative positional encodings which allow each word in the sentence to take into account its left and right context. The evaluation of the model is done on the TACRED dataset. 
The proposed model relies only on attention (no recurrent or convolutional layers are used), while improving performance w.r.t. the previous state of the art.
\end{abstract}

\section{Introduction}

Recently, much research has been dedicated to either supplementing RNN and CNN models with attention mechanisms or substituting them with attention-only approaches as proposed by \newcite{NIPS2017_7181} or \newcite{shen2018disan}. However, most of this research concentrates on the field of Neural Machine Translation (NMT). Until recently, a self-attention-only mechanism has not been widely used for other NLP tasks.

There are, however, approaches to slot filling that use attention mechanisms to improve the performance of an LSTM layer. An example of this is the approach using position-aware attention on top of LSTM proposed by \newcite{D17-1004}. In this research paper, we aim to combine the self-attention encoder formulated by \newcite{NIPS2017_7181} and augment it with the position-aware attention layer of \newcite{D17-1004}. Additionally, we propose various modifications to both of the approaches, most notably an attention weighting scheme that models pair-wise interactions between all tokens in the input sentence, taking into account their relative positions to each other.

\section{Background}

The task of relation classification can be paraphrased in the following way: \emph{Decide which relations (of a fixed set of given relations) hold between two selected entities in a sentence.}

The TAC KBP evaluations\footnote{https://tac.nist.gov/} provide a set of 42 frequent (pre-defined) relations for persons and organizations, and annotations as to whether those relations hold for selected entities in a sentence. Some examples are:
 \begin{itemize}
    \item \texttt{per:employee\_of}: Does (did) person X work for company Y?
    \item \texttt{org:city\_of\_headquarters}: Is (was) company/organization X based in city Y?
    \item \texttt{per:countries\_of\_residence}: Does (did) person X live in country Y?
    \item \texttt{per:title}: Does (did) person X have the job title Y?
 \end{itemize}

The TACRED dataset of \newcite{D17-1004} provides the TAC KBP data of the years 2009 to 2014 in a format that can be processed as a multiclass input-output mapping, which assigns each sentence (with relational arguments marked as Subject and Object) one of the relations of interest (or the special label \texttt{no\_relation}). An example instance is:\\
\textbf{input:} \emph{The last remaining assets of bankrupt Russian oil company $[$Yukos$]_{SUBJ}$ - including its headquarters in $[$Moscow$]_{OBJ}$ - were sold at auction for nearly 3.9 billion U.S. dollars on Friday .}\\
\textbf{output:} \verb=org:city_of_headquarters=

While the RNN-based architectures already include the relative and absolute relations between words due to their sequential nature, in the task of slot filling we not only need to take into account the sequence of words from start to end, but also to learn how the words relate to the query and the object in the sentence. 
The position-aware approach by \newcite{D17-1004} already models the interactions relative to the subject and object positions. However, interactions between all other words are only only dealt with by the LSTM layer. 

In our approach, we substitute the LSTM layer with the self-attention encoder, 
a mechanism that models all pair-wise interactions in an input sentence.
The self-attention approach itself does not model the sequential order of the input.
However, information about this order can be provided by embeddings of the (absolute) positions in the sentence,
and previous work
indicates that including relative positional representation in self-attention models improves performance for the task of Neural Machine Translation \cite{DBLP:journals/corr/abs-1803-02155}. Before we describe our approach to dealing with relative positional encodings in the self-attention encoder and also show how to combine the encoder with the position-aware attention layer, we provide more background on how the original implementation of these approaches work.

\subsection{Self-Attention Encoder (Google Transformer)}
\label{sect:transformer}

The Google Transformer model created by \newcite{NIPS2017_7181} is the first model that uses self-attention without any RNN or CNN based components. It is used for the task of Neural Machine Translation and has an encoder-decoder structure with multiple stacked layers. In the transformer model, the input representation for each position is used as a query to compute attention scores for all positions in the sequence. Those scores are then used to compute the weighted average of the input representations.

The attention is regarded as a mapping of query and key/value pairs to an output, each of which being represented by a vector. More specifically, a self attention layer provides an encoding for each position $i$ in the sequence, by taking a word representation at that position as the query (a matrix $Q$ holds the queries for all positions) and computing a compatibility score with  representations at all other positions (the values, represented by a matrix $V$). The compatibility scores w.r.t. a position $i$ are turned into an attention distribution over the entire sequence by the softmax function, which is used as a weighted average of representations $E$ at all positions, the resulting output representation for position $i$.

In multi-headed self-attention, input representations are first linearly mapped to lower-dimensional spaces, and the output vectors of several attention mechanisms (called \emph{heads}) are concatenated to form the output of one multi-headed self-attention layer. An encoder layer consists of a self-attention layer followed by a fully connected position-wise feed-forward layer.

For one attention head $a$ in the first self-attention layer, we obtain the vector for position $i$:
\begin{equation}
\vec h^{(a)}_i = \mbox{Attention}(W^{q(a)} \vec e_i, W^{K(a)} E, W^{V(a)} E)
\end{equation}
where $W^{q(a)}, W^{K(a)} , W^{V(a)}$ are linear transformations (matrices) to map the input representation into lower-dimensional space, and the function computing the resulting vector (from $\vec q=W^{q(a)} \vec e_i$, $K= W^{K(a)} E$ and $V=W^{V(a)} E$) is defined by:
\begin{equation}
\mbox{Attention}(\vec q,K,V) = V \mbox{softmax}(K^T \vec q)
\end{equation}


The self-attention architecture encodes positional information by adding sinusoids of various wave-lengths \cite{NIPS2017_7181} to the word representations.

\subsection{Argument Extraction with Self-attention}
\label{sect:argument_extraction}

To the best of our knowledge, the transformer model has not yet been applied to relation classification as defined above (as selecting a relation for two given entities in context).
It has however been applied as an encoding layer in the related setting of \emph{argument extraction} \cite{DBLP:journals/corr/abs-1803-01707}, which is similar to question answering (where the question is a pre-defined relation).

\newcite{DBLP:journals/corr/abs-1803-01707} apply various modifications to the original self-attention model by \newcite{NIPS2017_7181}, namely:

\begin{enumerate}  
\item The residual connection goes from the beginning of the self-attention block to the last normalization step after the feed-forward layer. In the original implementation, there are two residual connections within each layer.
\item Batch normalization \cite{pmlr-v37-ioffe15} is used instead of layer normalization \cite{DBLP:journals/corr/BaKH16}.
\end{enumerate}

In our experiments we observed improvements on the development data using this version rather than the original implementation by \newcite{NIPS2017_7181}. A more detailed overview of the results is given in the Subsection~\ref{sect:variations}.

\subsection{Position-aware Attention for Slot Filling}
\label{sect:pos-aware}

The position-aware attention approach to slot filling by \newcite{D17-1004}
uses an LSTM to encode the input and incorporates distributed representations of how words are positioned relative to the subject and the object in the sentence. This position-aware attention mechanism is used on top of a two-layer one directional LSTM network. In this implementation, the relative position encoding vectors are simultaneously computed relative to the subject and the object. To illustrate, if the sentence length is six words and the subject is at position two, the position encodings vector will take the following form: \([-1, 0, 1, 2, 3, 4]\), where position 0 indicates the subject. Later for each position, an embedding is learned with an embedding layer. The same applies to a separate vector denoting object positions, and effectively two position embedding vectors are produced, \(p^s = [p^s_1, ..., p^s_n]\) for subject embeddings and \(p^o = [p^o_1, ..., p^o_n]\) for object embeddings, both sharing a position embedding matrix P respectively \cite{D17-1004}.

The final computation of the model uses the LSTM's output state: a summary vector \(q = h_n\), the LSTM's output vector of hidden states \(h_i\), and the embeddings for the subject and object related positional vectors. For each hidden state \(h_i\) an attention weight \(a_i\) is calculated using the following two equations: 
\begin{equation} 
\label{eq:1}
 u_i = v^\mathsf{T} tanh (W_h h_i + W_q q + W_s p^s_i + W_o p^o_i)
\end{equation}

\begin{equation}
\label{eq:2}
 a_i = \frac{exp(u_i)}{\sum_{j=1}^{n} exp(u_j)}
\end{equation}
where \(W_h\) and \(W_q\) weights are learned parameters using LSTM while \(W_s\) and \(W_o\) weights are learned using the positional encoding embeddings. Afterwards, \(a_i\) is used to pass on the information on how much each word should contribute to the final sentence representation \(z\):
\begin{equation}
\label{eq:3}
 z = \sum_{i=1}^n a_i h_i
\end{equation}

This sentence representation is used in a fully-connected layer. Finally, a softmax layer is used to compute the most likely relation class \cite{D17-1004}.

\section{Proposed Approach}

The approach by \newcite{D17-1004} uses attention to summarize the encoded input instance. Here, input had been encoded using an LSTM, resulting in an hidden vector $h_i$ at any position $i$ in the sentence. ``Traditional attention'' is used, meaning that there is one sequence of weights, used for a weighted average of the hidden vectors.

We propose two main changes to this approach:

\begin{enumerate}
\item We replace the LSTM by a self-attention encoder that computes hidden vectors $h_i$ considering all pairwise interactions in the sentence (rather than sequential recurrences). Following the equations \ref{eq:1}, \ref{eq:2} and \ref{eq:3} from Subsection~\ref{sect:pos-aware}, instead of calculating \(W_h h_i\) and \(W_q q\) using LSTM, we extract them using the self-attention encoder. The vector of \(h_i\) values is a direct output of the encoder, while the summary \(q\) is extracted by a one-dimensional max pooling layer applied on the output vector.
\item We augment the self-attention mechanism with \emph{relative position encodings}, which facilitate taking into account different effects that are dependent on the relative position of two tokens w.r.t. each other.
\end{enumerate}


\subsection{Changes to Self-attention Encoder}

This subsection describes what aspects of the self-attention encoder we have changed, namely, a different training strategy, the structural changes and a different approach to positional encodings.

\subsubsection{Changes to Positional Encodings}
\label{sect:changes-to-pos-enc}

The self-attention layer proposed by \newcite{NIPS2017_7181}
does not directly model the sequential ordering of positions in the input sequence -- rather, 
this ordering is modeled indirectly, using absolute positional encodings with cosine and sine functions to encode each position as a wavelength.
Assuming that words in a text interact according to their relative positions (the negation \emph{``not''} negates a verb in its vicinity to the right) rather than according to their absolute positions (the negation \emph{``not''} negates a verb at position 12), modeling positional information burdens the model with the additional task of figuring out relative interactions from the absolute encodings.

Research by \newcite{DBLP:journals/corr/abs-1803-02155} shows in the context of machine translation that using relative positional encoding can improve the model performance. Here, we describe our approach to making positional encodings relative, and its application to relation classification.

Recall from Section \ref{sect:transformer} that one self-attention head $(a)$ computes a representation for a position $i$ from the weighted (and linearly transformed) input representations at all positions:

$$\vec h^{(a)}_i = V \mbox{softmax}(\vec z)$$

where $V$ is a matrix of the (linearly transformed) input vectors, and $\vec z$ contains the unnormalized weights for all positions. In the original self-attention model, $\vec z$ simply contains the pairwise interactions of the input representations:

$$z_{pair} = K^T\vec q$$

where $K$ is a matrix of the (transformed) input vectors, and $\vec q$ is the (transformed) input representation at position $i$ (for which one wants to compute $h_i$). For relative position weights with respect to position $i$, we compute a second score $z_{relpos}$, that interacts with relative position embeddings $\vec m_j$, stacked to form the matrix $M_i$:

$$M_i = \left[m_{1-i}, \dots, m_{-1}, m_{0}, m_{1}, \dots , m_{n-i} \right]$$

where $n$ is the length of the input sequence and the vectors $\vec m_j$ are parameters of the model (different $\vec m_j$ are learned independently for each attention head). The matrix $M_i$ arranges the relative position vectors exactly such that $m_0$ is at position $i$, and all other $m_j$ are ordered relative to that position. A query vector $r$ is computed analogously to $\vec q$ from the input at $i$: $\vec r = W^{r(a)}\vec e_i$.
The position score $z_{relpos}$ results from the interaction of $r$ with the relative position vectors in $M_i$:

$$z_{relpos} = M^T \vec r$$

Our final model uses both the pairwise interaction scores and the relative position scores by summing them together before normalization:

$$\vec h^{(a)}_i = V \mbox{softmax}(z_{pair} + z_{relpos})$$


\subsubsection{Low-level Design Choices and Training Setting}

\textbf{Self-attention Encoder Layer.} We take over the changes proposed by \newcite{DBLP:journals/corr/abs-1803-01707} which we described in Subsection \ref{sect:argument_extraction}, namely, we use Batch Normalization and only one residual connection. 
Furthermore, instead of initializing weights using Xavier \cite{pmlr-v9-glorot10a}, we use Kaiming weight initialization \cite{He:2015:DDR:2919332.2919814}. Also, instead of using ReLU \cite{Nair:2010:RLU:3104322.3104425} as an activation function, we use the randomized leaky rectified linear unit function, RReLU \cite{DBLP:journals/corr/XuWCL15}.

\textbf{Training.} In the implementation by \newcite{NIPS2017_7181}, the Adam optimizer \cite{Kingma2014AdamAM} with learning rate warm-up is used. In our approach we follow the learning strategy proposed by \newcite{D17-1004} with the following hyper-parameters: we use Stochastic Gradient Descent with a learning rate set to 0.1, after epoch 15 the learning rate is decreased with a decay rate of 0.9 and patience of one epoch if the $F_1$-score on the development set does not increase. All model variations are trained for 60 epochs.

\subsection{Changes to the Position-aware Attention Layer}
The attention-based position-aware relation classification layer encodes the relative positions w.r.t. the object and subject.
We make it easier for the model to capture this kind of information, by binning positions that are far away from the subject or object: The further away a word is from the subject or the object, the bigger the bin index into which it will fall is. For instance, if the length of the sentence is 10 words and the subject position is at index 1, a regular positional vector will take the following form: \([-1, 0, 1, 2, 3, 4, 5, 6, 7, 8]\). After introducing the relative position bins, the same position vector will change to: \([-1, 0, 1, 2, 3, 3, 4, 4, 4, 5]\).

\section{Experiments}

\subsection{Experimental Setup}
\label{sect:finalmodel}

In addition to introducing various structural changes to self-attention and position-aware attention, we also use a different set of hyper-parameters than those reported by \newcite{NIPS2017_7181} and \newcite{D17-1004}.

Instead of training word embeddings with a dimension of 512 as in the original implementation, we use a pre-trained GloVe word embedding vector \cite{pennington2014glove} with the embedding size of 300. Additionally, following the implementation of \newcite{D17-1004}, we append an embedding vector of size 30 for the named entity tags and an embedding vector of the same size for the part-of-speech tags, amounting to a final embedding vector size of 360. Moreover, we see an improvement in performance when adding object position embeddings to the word embeddings, which is done before the relative positional embeddings discussed in Subsection \ref{sect:changes-to-pos-enc} are applied in the self-attention encoder layer.

In the original self-attention encoder the implementation of the position-wise fully connected feed-forward layer uses the hidden size that is double the word embedding size. In our experiments, we see no direct improvement in either doubling or increasing the hidden size even more. However, lowering the hidden size contributes to a slightly better performance than when doubling it. In our implementation, the hidden size is half the size of the embedding vector, namely 130.

In the self-attention encoder instead of using a stack of six identical encoder layers, we use only one layer. Similarly, to the research of \newcite{DBLP:journals/corr/abs-1803-01707} where only two layers are used, we see no performance gain when using more than one layer. In fact, in the case of slot-filling, a decrease in performance when using more than one encoder layer is observed.

Additionally, using 3 heads in the Multi-Head Attention instead of 8 yields the best performance. Using more than 3 heads gradually degrades performance with each additional head.

We change our dropout usage compared to the one used by \newcite{NIPS2017_7181} where a dropout of 0.1 is used throughout the whole model. In our implementation, we use dropout of 0.4, apart from the Scaled Dot-Product Attention part of the self-attention encoder where we apply dropout of 0.1.

As described before, we train the model using Stochastic Gradient Descent with a learning rate of 0.1 and decay it using decay rate of 0.9 and epoch patience of one after epoch 15 if the performance on the development set does not improve. All models are trained for 60 epochs, and use a mini-batch size of 50.

\subsection{TACRED Evaluation}

The TACRED dataset \cite{D17-1004} used to evaluate the model consists of \textit{106.264} hand-annotated sentences denoting a query, object, and the relation between them. In addition to that, the dataset already includes part-of-speech tags as well as named entity tags for all words. The sentences that serve as samples are very long compared to the ones available in similar datasets, for instance, Semeval-2010 Task 8 \cite{hendrickx-EtAl:2010:SemEval}, with an average sentence length of 36.2 words. 

Furthermore, there often are multiple objects and relation types identified for each query within one sentence. Each query-argument relation example, however, is saved as a separate sentence sample. 

Moreover, 79.5\% of the whole dataset samples are query-argument pairs that do not have any relation between them and are labeled with a \textit{no\_relation} relation type. Overall, the dataset includes samples for 42 relation classes, out of which 25 are relations of type \textit{person:x} (i.e., \textit{person:date\_of\_birth}), 16 of type \textit{organization:x} (i.e., \textit{organization:headquarters}), and the \textit{no\_relation} class. 

The dataset is already pre-partitioned into \textit{train} (\textit{68124} samples), \textit{development} (\textit{22631} samples), and \textit{test} (\textit{15509} samples) sets. The dataset also comes with an evaluation script, which we use to run the subsequent evaluation. 

Table~\ref{model-comparison} shows the LSTM baseline results and the best model results reported by \newcite{D17-1004}, as well as our best model results for comparison. Our model exhibits better performance overall with a 1.4\% higher $F_1$-score than the state-of-the-art performance reported by \newcite{D17-1004}. While our model achieves lower precision, the recall is considerably higher with a 4.1\% difference. 

\begin{table}[h]
\begin{center}
\begin{tabular}{|l|r|r|r|}
\hline \bf Approaches & \bf P & \bf R & \(\mathbf{F_1}\) \\ \hline
LSTM & 65.7 & 59.9 & 62.7 \\
Position-aware LSTM & \bf 65.7 & 64.5 & 65.1 \\
Our model & 64.6 & \bf 68.6 & \bf 66.5 \\
\hline
\end{tabular}
\end{center}
\caption{\label{model-comparison} TACRED test set results, micro-averaged over instances.}
\end{table}

In addition to testing the single model results, we also follow the same ensembling strategy applied by \newcite{D17-1004}, where five models are trained with a different random seed and later on, using ensemble majority vote a relation class is selected for each sample. The comparison of the results is shown in Table~\ref{model-ensemble}. Our ensembled model reaches a slightly higher $F_1$-score as that of \newcite{D17-1004}, namely, 67.3\%. However, there are significant differences regarding precision and recall. Their ensembled model achieves a relatively high precision of 70.1\%, while our model reaches high recall of 69.7\%.

\begin{table}[h]
\begin{center}
\begin{tabular}{|l|r|r|r|}
\hline \bf Ensemble Models & \bf P & \bf R & \(\mathbf{F_1}\) \\ \hline
Position-aware model & \bf 70.1 & 64.6 & 67.2 \\
Our model & 65.1 & \bf 69.7 & \bf 67.3 \\
\hline
\end{tabular}
\end{center}
\caption{\label{model-ensemble} Comparison of ensemble models evaluated on TACRED test set, micro-averaged over instances.}
\end{table}

\subsection{Model Variations}
\label{sect:variations}

\begin{table}[h]
\begin{center}
\begin{tabular}{|l|r|r|r|}
\hline \bf Model Variation & \bf P & \bf R & \(\mathbf{F_1}\) \\ \hline
Lemmas & 64.3 & 66.9 & 65.6 \\
Default residual conn. & 61.7 & 69.7 & 65.4 \\
Layer normalization & 53.6 & 73.1 & 61.9 \\
ReLU instead of RReLU & 64.5 & 68.0 & 66.2 \\
LSTM with self-attention & 65.2 & 62.7 & 64.0 \\
Self-attention encoder & 26.7 & \bf 85.4 & 40.6 \\
Absolute pos. encodings & \bf 65.9 & 66.7 & 66.3 \\
Kaiming instead of Xavier & 64.3 & 68.8 & 66.5 \\
Final model & 64.6 & 68.6 & \bf 66.5 \\
\hline
\end{tabular}
\end{center}
\caption{\label{model-variations} Results of model variations evaluated on TACRED test set, micro-averaged over instances.}
\end{table}

In addition to the final model described in Subsection~\ref{sect:finalmodel}, we try various variations and modifications the results of which we report in this subsection. Table~\ref{model-variations} shows the results for the following variations:

{\bf Lemmas instead of raw words}: instead of using raw words we extract their lemmatized representations using the spaCy\footnote{https://github.com/explosion/spaCy} NLP toolkit. Using lemmas yields a small increase in precision but a lower recall. Overall, this approach achieves 65.6\% $F_1$-score.

{\bf Original residual connection in self-attention}: the model uses the default residual connections as described by \newcite{NIPS2017_7181}, namely, one residual connection is passed from before the multi-head attention into the normalization layer after it, and the second one going from before the feed-forward part to the next normalization layer. In this case, we see an overall high recall score of 69.7\% with a relatively low precision of 61.7\%. 

{\bf Layer normalization instead of batch normalization}: In the original self-attention implementation by \newcite{NIPS2017_7181}, layer normalization \cite{DBLP:journals/corr/BaKH16} is used. Here we show how model's performance changes when using it over batch normalization \cite{pmlr-v37-ioffe15}. By using layer normalization we can achieve a relatively high recall of 73.1\%, although the precision of 53.6\% is one of the lowest throughout all of our model variations with the exception of a variation using self-attention encoder without the position-aware layer.

{\bf Using ReLU instead of RReLU}: Using ReLU \cite{Nair:2010:RLU:3104322.3104425} as an activation function gives a 0.3\% lower $F_1$-score compared to the best model which uses RReLU \cite{DBLP:journals/corr/XuWCL15}.

{\bf Combining self-attention encoder with position-aware attention and LSTM}: In this model variation we use the LSTM hidden layer to compute the \(h_i\) in the equation \ref{eq:3} from Subsection~\ref{sect:pos-aware} while using self-attention encoder for the calculation of \(a_i\), as well as the underlying \(W_h h_i\) and \(W_q q\) in equations \ref{eq:1} and \ref{eq:2}. The final result does not yield any significant performance increase compared to other model variants.

{\bf Self-attention encoder without the position-aware attention layer}: We also test our model performance by only using the self-attention encoder without the position-aware attention layer. This is a particularly interesting experiment, since the model reaches the highest recall value of 85.4\% throughout all of our experiments, although at the same time achieving only 26.7\% precision.

{\bf Self-attention encoder without the relative positional encodings}: Using the original absolute positional encodings from the original self-attention encoder implementation also yields relatively good results compared to all of the other model variations. Overall, however, this approach is showing a 0.2\% lower $F_1$-score than when using the relative positional encodings.

{\bf Using Kaiming weight initialization instead of Xavier}: The original self-attention implementation uses Xavier \cite{pmlr-v9-glorot10a} weight initialization approach to initialize the weights for query, key and value matrices. In this comparison run, using Xavier over Kaiming weight initialization \cite{He:2015:DDR:2919332.2919814} exhibits the same $F_1$-score but increases the precision a bit. Ultimately, there is a very small difference between these initialization techniques, although both have a slightly different effect on precision and recall.

\section{Related Work}

While self-attention was originally used for the task of Neural Machine Translation, recently it was applied to other NLP tasks unrelated to NMT. 

\cite{Kitaev-2018-SelfAttentive} use the self-attention encoder instead of LSTM to improve a discriminative constituency parser and achieve state-of-the-art performance with their approach. 
\cite{j.2018generating} successfully use the self-attention decoder for the task of Neural Abstractive Summarization. There is also an ongoing research by OpenAI \footnote{https://blog.openai.com/language-unsupervised/} to use self-attention to pre-train a task agnostic language model.

Nevertheless, to the best of our knowledge, self-attention was not previously applied to the task of relation extraction. Apart from the position-aware attention for LSTMs by \newcite{D17-1004}, various other approaches exist. \newcite{Angeli14combiningdistant} uses pattern based extractor and a supervised logistic regression classifier for relation extraction. \newcite{DBLP:conf/naacl/NguyenG15} as well as \newcite{adelSF2016} use Convolutions Neural Networks. \newcite{Xu15classifyingrelations} use a modified LSTM architecture called SPD-LSTM.

\section{Conclusions}

In this work, we show that self-attention architecture can be effectively applied to relation classification, resulting in a model that is purely based on attention mechanism, and does not depend on other encoding mechanisms such as LSTM. In our experiments, using the self-attention encoder and combining it with a position-aware attention layer achieves better results on the TACRED dataset than previously reported by \newcite{D17-1004}. 

Additionally, we examine several changes to both of the approaches to make them more effective on the task of relation classification. The main change to the self-attention is that instead of using absolute positional encodings we successfully use relative positional encodings that increase the final performance. We also modify the way how the relative encodings in the position-aware layer are represented by grouping the word embeddings into bigger bins the further away they are from the subject or the object. 

By trying out various model variations, we can see that using self-attention encoder alone leads to a high recall value but a very low precision. As a result, using the position-aware layer with the self-attention encoder helps achieve stable results with precision and recall being very close to each other. 

As future work, we propose to investigate further variations of the self-attention encoder, and to do more research on why using multiple encoding layers and a higher number of heads does not improve the performance of the model. Moreover, since using only the self-attention encoder yields a relatively high recall value of 85.4\%, it is worth exploring other approaches to improving precision without compromising the high recall in this model variation.

\bibliographystyle{acl}
\bibliography{references}

\end{document}